\documentclass[preprint]{elsarticle}

\usepackage{amssymb}

\usepackage{graphicx} 
\usepackage[cmex10]{amsmath}
\usepackage[table]{xcolor}
\usepackage[caption=false,font=footnotesize]{subfig}
\usepackage{framed,multirow}
\usepackage{amssymb}
\usepackage{latexsym}
\usepackage{url}
\usepackage{booktabs}
\usepackage{xcolor} 
\usepackage{csquotes}
\usepackage{rotating,tabularx,ltxtable}
\usepackage{pgfplots}
\usepackage[flushleft]{threeparttable}
\usepackage{subfig}
\usepackage{natbib}
\usepackage{xspace}
\usepackage{multirow}
\usepackage{longtable}
\usepackage{diagbox}

\usepackage{tikz}
\usepackage{circuitikz} 
\usepackage{pgfplotstable}
\usepackage{tikz-3dplot} 
\usepackage{float}
\usepackage[outdir=./]{epstopdf}

\pgfplotsset{compat=1.18}

\newenvironment{red}{\color{red}}{}

\journal{Heliyon}

\begin{document}

\begin{frontmatter}

\title{A novel method for identifying rice seed purity  based on  hybrid  machine learning algorithms} 
\author[1]{Thi-Thu-Hong Phan \corref{cor1}} 
		\cortext[cor1]{Corresponding author:   hongptt11@fe.edu.vn }

\author[1]{Quoc-Trinh Vo} 
\ead{trinhvq@fe.edu.vn}
\author[2]{Huu-Du Nguyen} 
\ead{du.nguyenhuu@hust.edu.vn}

\affiliation[1]{organization={Artificial Intelligence Department,
FPT University},
           city={Da Nang},
           postcode={550000}, 
           country={Viet Nam}}

\affiliation[2]{organization={School of Applied Mathematics and Informatics, Hanoi University of Science and Technology},
            city={Hanoi},
            country={Vietnam}}
            

\title{}

\begin{abstract}
In the grain industry, the identification of seed purity is a crucial task as it is an important factor in evaluating the quality of seeds. For rice seeds, this property allows for the reduction of unexpected influences of other varieties on rice yield, nutrient composition, and price. However, in practice, they are often mixed with seeds from others. This study proposes a novel method for automatically identifying the rice seed purity of a certain rice variety based on hybrid machine learning algorithms. The main idea is to use deep learning architectures for extracting important features from the raw data and then use machine learning algorithms for classification. Several experiments are conducted following a practical implementation to evaluate the performance of the proposed model. The obtained results show that the novel method improves significantly the performance of existing methods. Thus, it can be applied to design effective identification systems for rice seed purity.
\end{abstract}



\end{frontmatter}


\section{Introduction}

Rice is one of the most important cereal grains for humans. It is a staple food full of nutrition and calories for over half of the humans in the world (\citet{soh2023automated}). 
As the global population is increasing day by day, the role of rice is becoming more important and it is important to increase substantially the current level of rice production.
An essential strategy for achieving high rice yield is to ensure the quality of rice seeds. A good seed has a high probability of producing a healthy, tolerant, and high-yielding rice plant. As a result, planting quality seeds is a necessary condition for a bountiful harvest, contributing to an increase in the quality of rice products.

There are several standards for the quality of rice seeds, including purity. The purity of seeds means that they must be genetically pure and free from all sorts of other seeds. This is an essential property of rice seeds to avoid unexpected influences of other varieties on their yield, nutrient composition, and price. However,  purebred rice seeds inevitably get mixed up with others due to several stages of planting, producing, packaging,  and circulating.  
The identification of rice seed purity
is therefore a crucial task for the grain industry.

Traditionally, this task is performed manually based on the experience of experts /technicians on the visual and physical characteristics of rice seeds. 
This method, however,  may give inaccurate and inconsistent identification results due to subjective factors such as the different experiences of experts and the fatigue from concentrating on the same job for a long time. Recently, advanced technologies like computer vision and machine learning have been applied to handle the mission of identifying the purity of rice seeds. \citet{hong2015comparative} used some machine learning (ML) algorithms including \textit{K}-nearest neighbor (KNN), support vector machine (SVM), and random forest (RF) to identify rice seeds of a variety from a set of images of rice seeds mixed with the images of other types. The authors also suggested using feature extraction techniques of computer vision like Global Image Descriptor (GIST) and scale-invariant feature transform (SIFT) to extract essential features from the images of rice seeds before feeding the  ML algorithms. 
Although this is a widely used strategy to improve the efficiency of ML models, it has certain limitations. Since the authors relied on some basic descriptors which are of the hand-crafted feature extraction method, several challenges could be faced. These challenges include identifying the relevant seed features capable of reliably discriminating the seed of interest, finding robust features concerning distortion, variations, and deformation in the environment, and finding invariant features concerning translation, rotation, and scale \citep{chaugule2016identification}. In addition, the performance of the method applied in \citet{hong2015comparative} may not be the one expected as the accuracy attained is less than 95\%. Therefore, it is necessary to design an automatic system and to improve the performance of the existing ML models for identifying rice seeds, meeting the classification requirements in practice. In this study, we suggest using a hybrid model combining deep learning (DL) networks and ML algorithms to take advantage of the two methods. That is, a deep convolutional neuron network (CNN) is used to extract crucial features from the dataset of rice seed images. These CNN features are then fed to the ML models to perform the classification task. We consider two different kinds of deep CNN, i.e. VGG16, and ResNet-50, and six different kinds of ML algorithms, including SVM, RF, KNN, Decision Tree (DT), Extra Trees (ET), and Logistic Regression (LR). These are common models that are widely used in the literature and have proven effective in many classification problems. In the use of VGG16 and ResNet-50, the features extracted from the several blocks inside the networks are applied rather than using the features extracted from the last block/layer as in many existing studies. Moreover, the experiments have been set up based on an implementation for rice seed identification in practice.  Details of the proposed algorithm, experimental setup, and obtained results will be presented in the next sections.


\section{Related works
\label{sec: related}}
Due to its importance in ensuring the yield and quality of rice planting, many studies have been devoted to identifying rice seed varieties in the literature.  
\citet{chaugule2014evaluation} evaluated the texture and shape features using the neural network architectures for the classification of paddy varieties. \citet{huang2017novel} utilized image segmentation and a key point identification algorithm to segment paddy seed images and extract seed features. These features were then fed to a back propagation neural network for identifying three kinds of paddy seed varieties in Taiwan that are indistinguishable by inspectors during seed purity inspections. The back propagation neural network combined with the image processing method was also applied in \citep{abbaspour2020combined}  to classify 13 indigenous rice cultivars of Iran.  A review of machine vision systems for food grain quality evaluation was conducted in \citep{vithu2016machine}. In general, recent studies tend to take advantage of advanced techniques of artificial intelligence and computer vision in designing rice seed identification models. A common scheme can be seen in these studies. Firstly, they rely on hand-crafted feature extraction approaches to extract important features from the rice seed images. The most widely used features include color, shape, morphology, and texture. Then, the principle component analysis technique is performed to reduce data dimensionality and visualize the underlying structure in experimental data. After the reduction, the processed data are fed to the ML models for the identification of rice seed purity. Typical studies that followed this scheme can be seen in \citet{kuo2016identifying,anami2019automated} and \citet{ansari2021inspection}. A drawback of the approach is that it requires a significant effort for the hand-crafted extraction of important features.  In the literature, there is also another scheme that exploits the power of hyperspectral images in identifying rice seeds. The spectral features extracted from the hyperspectral images are processed using several computer vision techniques and then put into machine learning or deep learning models. Typical studies of this scheme are presented in {\red \citet{ bai2020rapid, zhang2023hyperspectral, zhang2019purity}  and \citet{tu2022model}}. Following this scheme, it is necessary to have a modern imaging spectrograph, a high-performance camera, and a compatible lens. However, it should be considered that these tools are not very popular, only in some well-equipped labs or even grain companies. 

When it comes to the classifier, \citet{liu2021rice} split the rice seed identification methods into two categories, the popular deep learning algorithms for classification and recognition, and the traditional machine learning methods for identification. Each has its advantages and disadvantages. While feature extraction from raw data is not necessary for deep learning, it asks for a large number of training samples,
leading to a high cost of data collection. By contrast,  machine learning methods do not require many training samples to achieve a satisfactory identification performance, but a significant effort for reducing redundant information
and unpredictable noise from raw data. \citet{anami2020deep} compared the performances of three different machine models, namely, Multilayer Back Propagation Neural Network (BPNN), SVM, and KNN using the color and texture features of rice images extracted from the Principal Component Analysis (PCA) and Sequential Forward Floating Selection (SFFS) methods. 
However, the best performance of these three methods is only about 93\%. Aiming for better performance,  researchers have recently paid more attention to using CNN-based transfer learning models and achieving effective results even with a small number of training samples. For instance, the VGG16 network has been applied for rice seed classification and demonstrated good performance in many studies such as \citet{panmuang_image_2021,sun2022enhanced} and \citet{de2021irrigated}. Other pre-trained models such as Inception-V3, VGG-16, ResNet-50, DenseNet-121, and MobileNet-28 have also been used to identify and classify paddy crop biotic stresses from the field images in   \citet{malvade2022comparative}. \citet{eryigit2021performance} investigated the efficiency of these models and their deeper variants like VGG-19, Resnet101,  Resnet152, DenseNet201, and EfficientNet in classifying seven different grass species. A comparative study of rice variety
classification based on several deep learning architectures, including AlexNet, VGG, Inception, ResNet, DenseNet, MobileNet, and NASNet has been conducted in \citet{hoang2020comparative}. The authors stated that all CNN models
provided very good results on the rice seed classification task. However, the setup of their experiment is not very popular as the practitioners/engineers rarely perform the task of identifying multiple varieties from experimental samples to determine the purity. A more practical setup for the experiments identifying rice seed purity is therefore performed in this study. {\red The same idea of the setup applied for the genuineness detection
of maize seeds was also considered in \citet{tu2021non}.}

\section{Data collection}
\begin{figure}[!th]
\begin{center}

\includegraphics[width=0.98\linewidth]{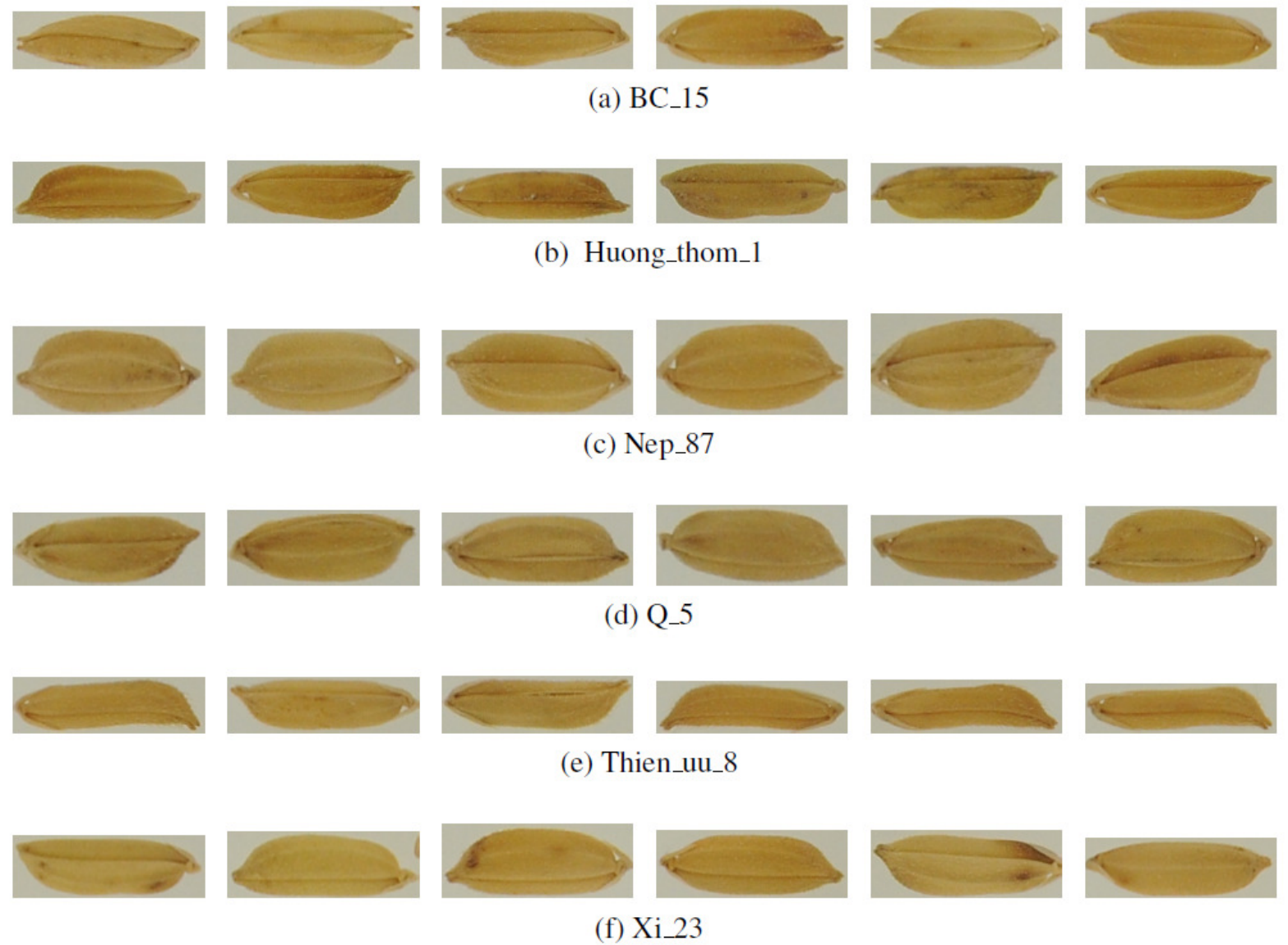}

\end{center}
\caption{An example of the rice seed images from the dataset}
\label{fig:riceImages}
\end{figure}

In this study, a dataset consisting of images of six Vietnamese rice varieties, namely, BC\_15, Huong\_thom\_1, Nep\_87, Q\_5, Thien\_uu\_8, and Xi\_23 is considered. These are the most six common rice varieties cultivated in the north of Vietnam. 
The rice seeds were collected carefully from a rice seed company under standard conditions for rice seed production. All the seeds were packed and labeled separately. In particular, about 300 seeds were picked up each time from each variety and spread evenly on a table. Then, the pictures of these seeds were taken using a CMOS image sensor color camera (NIKON D300S) following strict technical conditions to ensure their quality. The camera was fixed at a distance of 400 mm from the plane where the rice samples were placed. LED light panels measuring 300x300 mm with a color temperature of 5000 K and a 14 W light source are arranged around, with one area serving as a photography background. The system is equipped with dark room shooting and remote control, avoiding vibration and ensuring image quality.
After that, the image of each seed was cropped individually and gathered in the same folder,  preparing for the experiments. Figure \ref{fig:riceImages}  provides an example of the rice seed images from the dataset. 

\begin{figure*}[!th]
\begin{center}
\includegraphics[width=0.98\linewidth]{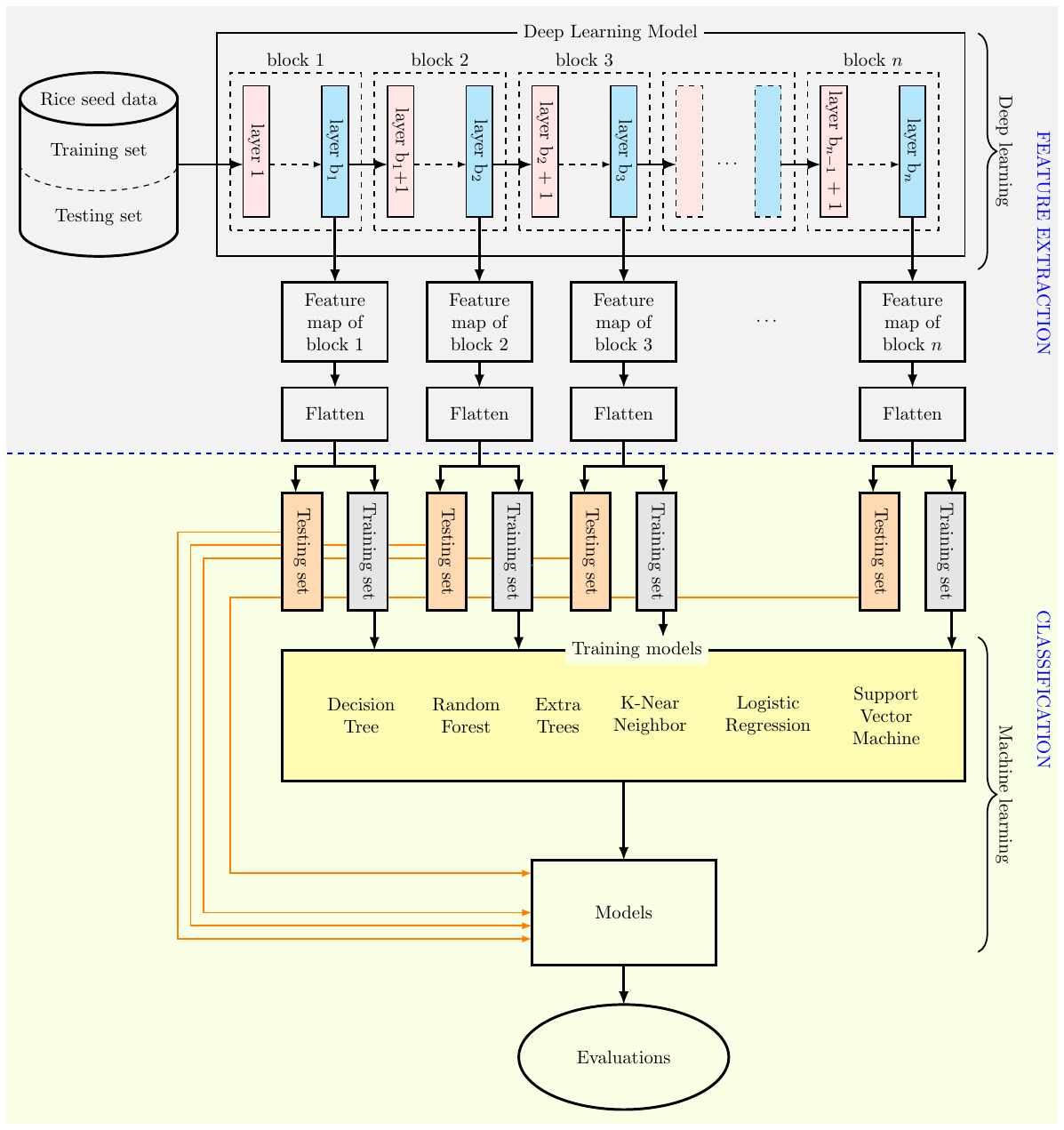}
\end{center}
\caption{The schema of the proposed method }
\label{fig:methodology}
\end{figure*}

\section{The proposed method}
The main idea of the proposed method is to use deep CNN algorithms for extracting essential features from the rice seed images and to use popular machine learning algorithms for classifying the rice seeds based on these features. That is, it includes two main steps: feature extraction and classification as visualized in Figure \ref{fig:methodology}. Feature extraction is a very important technique in machine learning. It allows reducing significantly the amount of redundant data from the dataset without losing vital information. It also plays a key role in improving the efficiency and accuracy of machine learning models. Instead of using hand-crafted methods as in many existing studies in the literature,  this study proposes to take advantage of deep CNN models to automatically extract features of rice seed images. Since a deep CNN model contains several blocks, we aim to apply the features extracted from not only the final blocks but also the other blocks inside the networks. These features are then sent to the machine learning models to train the models and to perform the identification task. 

\subsection{Deep learning methods for feature extraction} 
 CNN is a major class of deep neural networks that are widely used in computer vision. In this section,  two architectures applied in the study, namely  VGG 16 and ResNet-50, will be briefly presented.
 
\subsubsection{VGG16}
\begin{figure*}[!th]
\begin{center}
\includegraphics[width=0.98\linewidth]{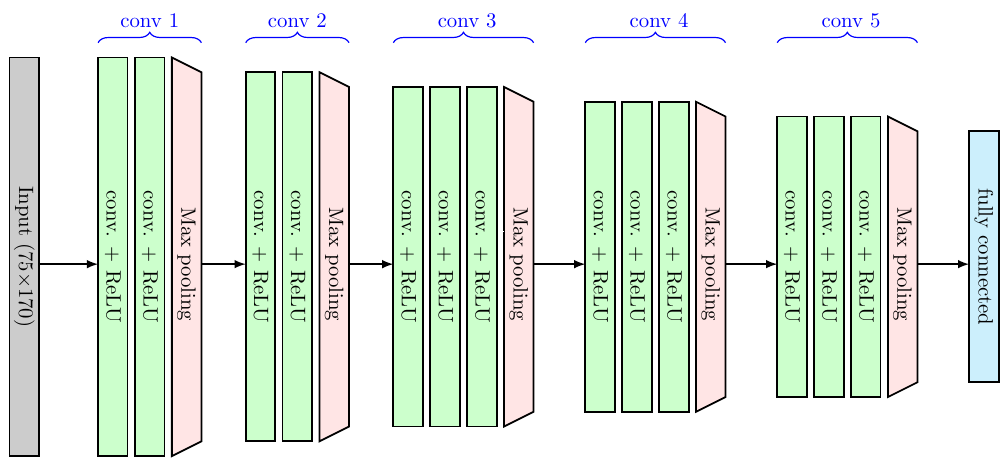}
\end{center}
\caption{The architecture of the VGG16 model: Convolutional layer with kernel of 3$\times$3 and max pooling of 2$\times$2 }
\label{fig:vgg16Arch}
\end{figure*}

VGG16 refers to the VGG model, a CNN model containing 13  convolutions and
pooling layers followed by 3 fully connected layers. The model was proposed and published  in \citep{simonyan2014very}    by
researchers at the Visual Geometry Group at the University
of Oxford.   It achieved an accuracy of 92.3\%, which is in the top-5 of the ImageNet Large Scale Visual Recognition Challenge in 2014. The VGG16 architecture comprises five main blocks.  The first two blocks include two convolutional layers, each followed by Max pooling layers, while the last three blocks contain three convolutional layers, each followed by   Max pooling layers as shown in Figure \ref{fig:vgg16Arch}. The input image size is fixed at $75 \times 170$, which is fit with the image of each seed. After pre-processing, they were passed through a convolutional layer with a $3 \times 3$ filter with a stride of 1. The Max pooling layers are performed with $2 \times 2$  sizes and stride is set to 2. Three fully connected layers have the same configuration of 4096 neurons and ReLU activation in each layer.

Thanks to the structure of multiple blocks with alternating convolutional and pooling layers, VGG16 can learn effectively complex and abstract representations from the input. The pooling layers in each block help to reduce the spatial dimensions of the feature maps, making the representation more robust to translations and decreasing the computational burden for the next layers. Also, the use of small convolutional filters ($3 \times 3$)  allows it to capture fine-grained details of the input, while the deep architecture results in the capacity to model large amounts of data variation. As a result, VGG16 has been broadly used for different tasks such as image classification, object detection, and segmentation. 

\subsubsection{ResNet-50}

In deep learning, a strategy to improve the accuracy of a  model is to increase the number of layers. However, it also leads to the problem of a vanishing gradient that prevents the model from converging. The Residual Network (ResNet),  overcomes this difficulty by using the gradient clipping technique for updating the error derivative. In particular, it utilizes skip connections that allow inputs to “skip” some convolutional layers.  After learning a given feature, it focuses on learning new features rather than attempting to learn it again. This makes ResNet much deeper than previous architectures while still achieving high accuracy. The network won the ImageNet Large Scale Visual Recognition Challenge (ILSVRC) in 2016, achieving a top-5 accuracy of 95.5\%.

\begin{figure*}[!th]
\begin{center}
\includegraphics[width=0.98\linewidth]{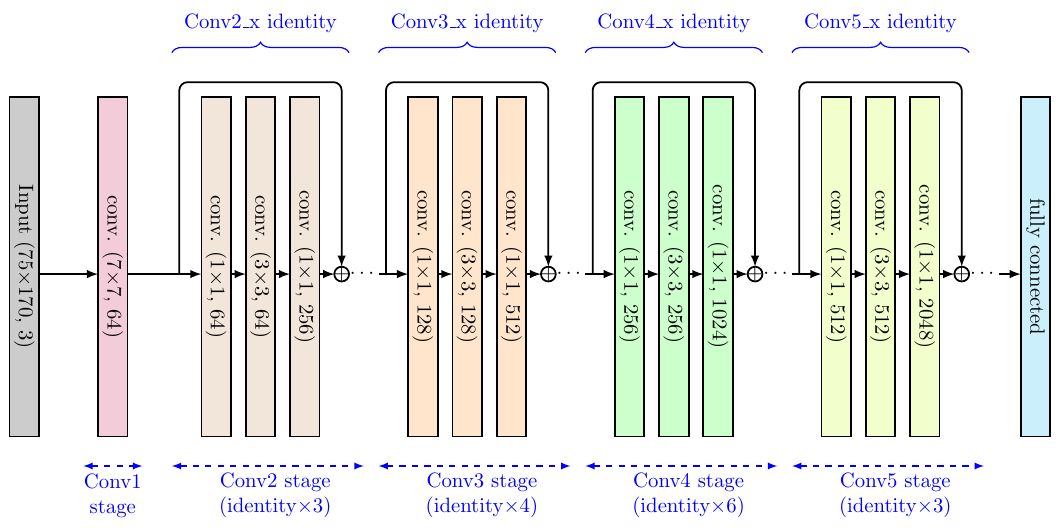}
\caption{The architecture of the ResNet-50 model }
\label{fig:resnet50Arch}
\end{center}
\end{figure*}

The model comprises a total of 50 layers, including convolutional and fully connected layers, and uses shortcuts to bypass one or more layers. These layers are divided into five main stages as depicted in Figure \ref{fig:resnet50Arch}. 

The first stage is a stem layer with a 7$\times$7 convolutional layer followed by a 3x3 max pooling layer, both with a stride of 2. The next four stages are composed of several (depending on the stage) residual blocks, each containing multiple convolutional layers. A typical residual block consists of three consecutive layers, i.e. a $1\times 1$ convolution layer, a 3$\times$3 convolution layer, and another 1$\times$1 convolution layer. The role of the 1$\times$1 convolutions is to reduce and then increase the dimensions of the input, while the 3$\times$3 convolution is for extracting features.  
In each block, the final output is the combination of two parts, the input from the previous block and the output of three convolutional layers. The residual blocks are then followed by a global average pooling with the function of averaging the spatial dimensions of the feature maps. A fully connected layer comprises 1000 neurons, corresponding to the 1000 classes in the ImageNet dataset. The detailed structure of the CNN  model was first introduced in \cite{he2016deep}. 


\subsection{Machine learning algorithms}

\subsubsection{Decision Tree, DT}

Decision Tree (DT) is one of the most classical algorithms of machine learning that is used for solving both classification and regression problems. The algorithm aims to predict the value/class of a dependent variable by learning simple decision rules inferred from the features of the data. It splits the dataset into 2 branches based on the binary split process at each node. This separation procedure is then applied to develop new branches until each branch becomes inseparable. 
 

\subsubsection{Random Forest, RF}
Random Forest (RF) refers to an ensemble of decision trees, as it is made up of multiple single trees that work together to give the final result. Each tree in the forest is trained on a bootstrap set of the data and a  random subset of the attributes is selected at e for splitting at each node. This makes the trees in RF more diverse,  and less correlated.  Furthermore, in the building of RF, fully grown trees without pruning are considered to keep the low bias. 
 As the number of trees in the algorithm has a significant impact on its efficiency,  it is crucial to determine this value for a concrete application.  A discussion for the choice of the number of trees in the RF can be seen in \citet{oshiro2012many}.



\subsubsection{Extra Trees, ET}

Extremely Randomized Trees, or Extra Trees (ET) for short,  can be considered an extension of the RF that includes numerous un-pruned decision trees. ET aims to increase the randomness in the process of feature selection by randomly choosing features to split at each node. At each node of the RF, it chooses the best feature to split,  while in ET a random subset of features is selected and the best feature of this subset is taken to split. This makes the trees in ET  more diverse than those in RF  and decreases the correlation between trees. 

In the ET, the final prediction is determined based on the majority voting or the average value of the decision trees.  Several important parameters need to be determined, including the number of decision trees, the number of random features used for each split, and the minimum number of samples required at a node to create a new split point. Theoretically, the more the number of decision trees in ET, the better capacity it can result in. However, it also requires more time for the calculation while the obtained result may be negatively affected once it is beyond a critical number of trees.

\subsubsection{K-Nearest Neighbors, KNN}
  K-Nearest Neighbors (KNN) is an instance-based learning algorithm. It has been also known as a non-parametric statistical learning method since the classification predictions could be done without the requirement of data distribution.  The algorithm computes the distance between the new data point and all the other data points in the training set based on ``K" nearest neighbors and makes classification predictions considering the majority class label of these neighbors.



\subsubsection{Logistic Regression, LR}

Logistic Regression (LR) is a statistical method proposed to solve binary classification problems where the result is either a "yes" or "no" decision. This method uses the logistic function to model the probability of the outcome class.  The outcome depends on the conditional probability $ P(y|x) $ of a label $y$ given observation $x$ and the ratio $\dfrac{P(y=1|x)}{P(y=0|x)}$ is called the odds of an instance $x$ in the dataset. If the ratio is greater than 1, then x is assigned to label $y=1$. Otherwise, it is assigned to label $y=0$. Using an exponential of a linear function for approximating the odds of $x$ is the main idea of LR. 



\subsubsection{Support Vector Machine}

Support Vector Machine (SVM) is a kind of supervised learning algorithm in which the algorithm aims to find the optimal hyperplane that maximizes the margin between the closest data points of each class, also known as support vectors. It can deal with both linear and non-linear relationships between the predictor variables and the target variable. In the context of linear relationships, the optimal hyperplane is a simple line or plane, that can be expressed by  
    \begin{equation*}
        f(x)= {w}^Tx + b, 
    \end{equation*}
where  $w$ is the weight vector, $b$ is the bias and $x$ is the input vector. In the context of non-linear relationships, by applying Kernel functions, the original data will be transformed into a new space that has higher dimensions where a linear hyperplane can be used to separate the classes. 


\section{Experiments and results \label{sec: results}}
In this section,  we first describe the experimental setting and then discuss the performance of the proposed identification method for rice seed purity from these experiments.  The experiments are conducted using Google Colab Pro powered by the Tesla T4 GPU with Persistence-M 15G diver of version 510.47.03. This configuration provides a high-performance GPU computing environment for performing several different applications of machine learning and deep learning algorithms.

\subsection{Experiment setting}
\begin{figure}[!th]
\centering
\includegraphics[width=0.75\textwidth]{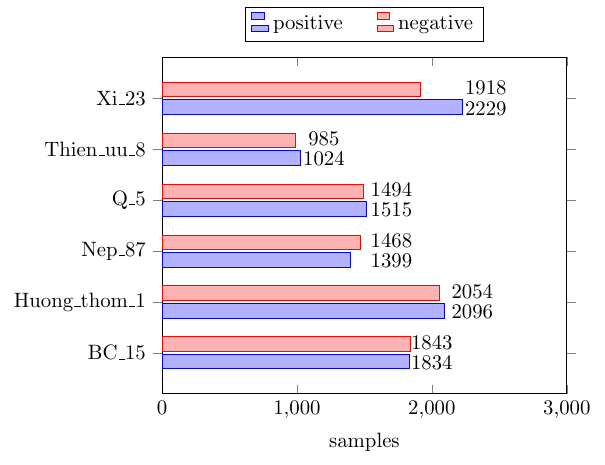}
\caption{The number of positive and negative samples in each folder of the dataset. \label{fignew}}
\end{figure}

The training set and testing set for the experiments are created as follows. Firstly, from six folders that contain the labeled images of six rice seed varieties, several images in each folder have been chosen and labeled as positive samples. Then, the images from another five folders were randomly picked up and labeled as negative ones. The number of negative samples is chosen so that it is almost equal to the number of positive ones. These images are mixed and put in a new folder labeled by the name of the rice seeds of positive samples. Accordingly, six folders corresponding to six datasets are created each containing images of rice seeds from six varieties labeled as positive and negative samples. Fig. \ref{fignew} displays the number of positive and negative samples in each folder. 

The function of the model now is to recognize only the variety of interests from others in each dataset. It should be considered that this setting follows a practical process and is different from the setting in many other studies in the literature. Indeed, the existing studies on rice seed identification often mixed the images from several varieties and used machine learning models to classify all these varieties. In other words, they aimed to perform a multi-class classification. However, the work of identifying the rice seed purity of a variety in practice is to check if there are seeds of different types, no matter which variety, mixed with seeds of the one of interest. That is, the task is to perform a binary classification. The random mix of images from all the varieties is also necessary since we can not know which variety of seeds could be mixed. The setting in our experiments was consulted by experts and towards the implementation in practice.  In the literature, the problem of transforming variety multi-classification into binary classification has also been considered in some situations, see, for example, \citet{lei2021docc,ali2020machine}.

To ensure a fair comparison of different classification methods, we fix the test set and training set and apply the out-of-bag technique suggested in \citet{Breiman_2001_RF} for estimating the generalization error. That is, each dataset is randomly split further into 
the training set and the testing set in the ratio of 67\% and 33\% as presented in Table \ref{rice_name}.

\begin{table}[!th]

\caption{Description of rice seed image dataset used for machine learning algorithms \label{rice_name}}

\begin{center}

\begin{tabular}{@{}llll@{}}
\toprule
Rice variety   & Training set & Testing set & Total \\ \midrule
BC\_15           & 2462         & 1215        & 3677  \\
Huong\_thom\_1 & 2780         & 1370        & 4150  \\
Nep\_87          & 1920         & 947         & 2867  \\
Q\_5             & 2015         & 994        & 3009  \\
Thien\_uu\_8    & 1345         & 664         & 2009  \\
Xi\_23           & 2778         & 1369        & 4147  \\ \bottomrule 
\end{tabular}
\end{center}
\end{table}

The efficiency of the proposed method  is measured using the metric of accuracy calculated by the formula:
\begin{equation*}
   \text{Accuracy} = \dfrac{TP+TN}{TP+TN+FP+FN} 
\end{equation*}
where TN (True Negative) is the number of negative samples classified accurately; TP (True Positive) is the number of positive samples classified accurately; FP (False Positive) indicates the number of actual negative samples classified as positive; and FN (False Negative) represents the number of actual positive examples classified as negative. Accuracy is one of the most commonly used metrics while performing classification. By its definition, this metric shows the proportion of correctly
classified rice seeds.


\subsection{The performance of deep learning methods}
To evaluate the performance of the proposed hybrid method, we first conduct experiments using deep convolutional neuron networks only. That is, the training set is used to train VGG16 and ResNet-50 networks, and then the efficiency of the trained models has been evaluated in the testing set. In the experiments, each model is trained with 100 epochs in total.  
To reduce training time, a transfer learning technique is applied while using VGG16 and ResNet-50 models.
Technically, all layers of these two models are frozen during the first 10 epochs to make sure the information they contain is not destroyed during further training rounds.
Then, several trainable layers are added on top of the frozen layers to turn the learned features into predictions.
Finally, fine-tuning is applied to training the whole model with the last 90 epochs to find the best model for each variety.

Table \ref{tab:DL_perf} shows the accuracy of these two deep CNN models in identifying rice seeds. As can be seen from the table, the obtained accuracy is quite low. With VGG, there are 3 rice varieties with an identification ability of only about 50\%. The accuracy of ResNet-50 is a bit better, but also only about 90\%. This means that the pure use of deep CNN models is not an effective way for the task of rice seed purity identification.

\begin{table}[!th]

\caption{The accuracy (\%)  of transfer learning methods in identifying rice seed purity with VGG16 and ResNet-50 \label{tab:DL_perf} }

\begin{center}
\begin{tabular}{@{}ccccccc@{}}
\toprule
DL networks   & BC\_15  & Huong\_thom\_1& Nep\_87 &Q\_5  &Thien\_uu\_8  &Xi\_23 \\ \midrule
VGG16&49.88&88.03&51.21&{ 49.70}&94.43&53.76\\
ResNet-50&84.36&{ 92.70}&89.76&69.72&89.16&46.38 \\
\hline
\end{tabular}
\end{center}
\end{table}

\subsection{The performance of proposed method using  VGG16 network}
Due to the use of convolutional layers with a stride greater than 1 and/or pooling layers in a deep CNN  model to extract features from data, when the input is passed through the network, the amount of features is reduced from layer to layer. Previous studies tend to take the output extracted from the last layers which are right before the classified layer. This allows the selection of notably compact data while including the feature that is needed for the classifier. However, this also means some other important features can be inadvertently omitted. In this study, we suggest considering the features extracted from different blocks in VGG16 rather than using the features extracted from the last block only. 

As mentioned above, a VGG16 network includes five main blocks.  Thus, one may have five choices to take features extracted from each block. However, in practice, the number of features extracted from block 1 and block 2 of the network is quite large, leading to a significant resource for storing and computing. We, therefore, take the features extracted from the last three blocks only. Tables \ref{VGG_ML_block3} - \ref{VGG_ML_block5} present the accuracy of the proposed method using the output from these blocks. Some important remarks from the obtained results can be drawn as follows.

\begin{itemize}
    \item In general, the proposed hybrid method using the VGG16 network results in a better performance for the rice seed identification compared to the use of either a deep learning model or a machine learning model. Indeed, 
    the use of ML models or VGG16 to classify rice seeds directly leads to poor performance as can be seen in Table 4 in \citet{hong2015comparative} and Table   \ref{tab:DL_perf} in this study. Meanwhile, the proposed hybrid method with the use of features extracted from block 3 of  VGG16 and the LR or SVM algorithms provides an accuracy of greater than 95\% for the identification of all the rice seed varieties, see Table  \ref{VGG_ML_block3}.
   
    \item The performance of the ML algorithms decreased significantly from Table \ref{VGG_ML_block3} to Table \ref{VGG_ML_block5} for all the cases. For example, with the BC, the DT algorithm provides an accuracy of 82.88\% in Table \ref{VGG_ML_block3}, which is greater than the corresponding values of 80.74\% and 76.63\% in Table \ref{VGG_ML_block4} and  Table \ref{VGG_ML_block5}, respectively. Thus, it can be concluded that the features extracted from the previous blocks are more likely to represent the data better than the following ones, leading to better rice seed identification for ML algorithms.
   
    \item The reason for the use of several ML models in this study is to find out the most effective models in recognizing rice seeds from different varieties. 
    Among these ML models, LR and SVM algorithms outperform the others. For example, the accuracy of the LR and SVM in identifying Q5 from other varieties in Table \ref{VGG_ML_block3} is both 95.07\%, remarkably higher than the accuracy of other ones which are all less than 84\%. This result can be explained by the fact that the task of rice seed purity identification is a binary classification problem. It aims to recognize the seeds from a rice variety of interest and to check whether or not this rice variety is mixed with other seeds.  LR and SVM are two specialized algorithms for binary classification, as a result, they provide the best performance in the experiments.  
    Between the two methods, the LR algorithm requires less time to train the model. For example, it took about 285 minutes to train SVM with Xi, which is approximately 9 times higher than the time training LR with the same variety.

\end{itemize}

\begin{table}[!th]

\caption{The {\red binary classification accuracy} (\%)  of proposed methods based on features extracted from VGG16  block 3. The best results are in bold. \label{VGG_ML_block3}}
\begin{center}
\begin{tabular}{@{}lrrrrrr@{}}
\toprule
Variety   & \multicolumn{1}{l}{DT} & \multicolumn{1}{l}{ET} & \multicolumn{1}{l}{RF} & \multicolumn{1}{l}{KNN} & \multicolumn{1}{l}{LR} & \multicolumn{1}{l}{SVM} \\ \midrule
BC\_15         & 82.88                  & 91.69                   & 93.00                     & 85.27                    & \textbf{97.20}                  & 97.04                   \\
Huong\_thom\_1 & 83.87                   & 92.56                   & 94.53                   & 87.81                    & 98.83                   & \textbf{99.20}                    \\
Nep\_87        & 88.17                   & 95.25                  & 94.72                   & 84.06                      & \textbf{97.89}                   & \textbf{97.89}                    \\
Q\_5         & 73.24                   & 80.38                   & 83.30                   & 77.77                    & \textbf{95.07}                   & \textbf{95.07}                   \\
Thien\_uu\_8   & 86.60                   & 97.14                   & 97.59                   & 94.43                    & \textbf{98.80}                   & 98.65                   \\
Xi\_23         & 82.54                   & 90.94                   & 94.38                   & 83.57                    & 96.64                  & \textbf{96.86}                  \\ \bottomrule
\end{tabular}
\end{center}
\end{table}

\begin{table}[!th]

\caption{The {\red binary classification accuracy} (\%)   of proposed method based on features extracted from VGG16  block 4. The best results are in bold. \label{VGG_ML_block4}}
\begin{center}

\begin{tabular}{@{}lrrrrrr@{}}
\toprule
Variety                          & \multicolumn{1}{l}{DT}      & \multicolumn{1}{l}{ET}      & \multicolumn{1}{l}{RF}      & \multicolumn{1}{l}{KNN}     & \multicolumn{1}{l}{LR}      & \multicolumn{1}{l}{SVM}      \\ \midrule
BC\_15                                & 80.74                        & 90.45                        & 93.33                        & 79.26                        & 96.54                        & \textbf{96.63}                        \\
Huong\_thom\_1 & 83.21 & 94.38 & 95.7 & 81.90 & 98.25 & \textbf{98.47} \\
Nep\_87                               & 87.33                        & 95.14                        & 94.83                        & 81.73                        & 98.20                        & \textbf{98.21}                        \\
Q\_5                                & 76.66                        & 80.89                        & 82.09                          & 74.75                        & \textbf{94.57}                        & 94.47                        \\
Thien\_uu\_8                          & 91.27                        & 98.04                          & 98.19                        & 93.83                        & 98.80                        & \textbf{98.95}                        \\
Xi\_23                                & 83.05                          & 92.40                        & 93.50                        & 80.79                        & 96.79                        & \textbf{96.93}                        \\ \bottomrule
\end{tabular}
\end{center}
\end{table}

\begin{table}[!th]

\caption{The {\red binary classification accuracy} (\%)   of proposed method based on features extracted from VGG16  block 5. The best results are in bold. \label{VGG_ML_block5}}
\begin{center}

\begin{tabular}{@{}lrrrrrr@{}}
\toprule
Variety                         & \multicolumn{1}{l}{DT}      & \multicolumn{1}{l}{ET}      & \multicolumn{1}{l}{RF}      & \multicolumn{1}{l}{KNN}     & \multicolumn{1}{l}{LR}    & \multicolumn{1}{l}{SVM}      \\ \midrule
BC\_15                                & 76.63                        & 86.34                        & 87.73                        & 77.94                        & \textbf{92.43}                      & 92.18                        \\
Huong\_thom\_1 &  79.27 & 90.21 &  92.26 &  77.15 & \textbf{94.01} &  93.94 \\
Nep\_87                               & 83.21                        & 92.40                        & 92.82                        & 86.27                        & \textbf{95.14}                      & 94.83                        \\
Q\_5                                & 69.42                        & 77.67                        & 79.28                        & 73.64                        & \textbf{86.92}                     & 85.31                        \\
Thien\_uu\_8                          & 88.40                        & 96.69                        & 96.54                        & 89.46                        & \textbf{97.29}                      & \textbf{97.29}                        \\
Xi\_23                                & 72.10                          & 85.54                        & 87.66                        & 78.67                        & \textbf{94.08}                        & 93.21                        \\ \bottomrule
\end{tabular}
\end{center}
\end{table}

\subsection{The performance of proposed method using ResNet-50 network}

Similar to the use of VGG16, when using  ResNet-50 we only use features extracted from Stage 5 to avoid the excessive amount of features outputted from Stage 1 to Stage 4. Three sets of features extracted from three blocks in Stage 5 of ResNet-50 are fed to the ML algorithms for the rice seed classification. The obtained results are presented in Tables \ref{Resnet50_ML_block1} - \ref{Resnet50_ML_block3} and the same results as the previous experiment with the VGG16 network are observed. This is shown through the following comments.

\begin{itemize}
    \item The proposed model combining ResNet-50 with the ML algorithms performs better than the model based purely on using ResNet-50. For example, the use of ResNet-50 with SVM for Xi\_23 provides an accuracy of 95.76\% (Table \ref{Resnet50_ML_block1}), which is more than twice the accuracy of 46.38\% from the use of ResNet-50 only (Table \ref{tab:DL_perf}). 
    \item In addition, the degree of accuracy shown in Table \ref{Resnet50_ML_block1} is higher than the ones in Table \ref{Resnet50_ML_block2} and Table \ref{Resnet50_ML_block3}, meaning that the features extracted from the previous blocks of Stage 5 are more effective than the ones followed. 
    \item Among several ML algorithms, LR and SVM are still the best. Take the case of Thien\_uu\_8  with features extracted from block 1 in Table \ref{Resnet50_ML_block1} as an example. While LR and SVM both give an accuracy of 95.6\%, other algorithms produce an accuracy of less than 90\%. Finally,  the use of features extracted from Stage 5 of the ResNet-50 network is slightly less effective compared to the use of the ones extracted from Block 3 of the VGG16 network. This is demonstrated by the smaller values in Table \ref{Resnet50_ML_block1} compared to the corresponding values in Table \ref{VGG_ML_block3}. 
\end{itemize}

\begin{table}[!th]
\caption{The {\red binary classification accuracy} (\%)   of proposed method based on features extracted from ResNet-50  block 1 of Stage 5. The best results are in bold.\label{Resnet50_ML_block1}}

\begin{center}
\begin{tabular}{@{}lrrrrrr@{}}
\toprule
Variety& \multicolumn{1}{l}{DT} & \multicolumn{1}{l}{ET} & \multicolumn{1}{l}{RF} & \multicolumn{1}{l}{KNN} & \multicolumn{1}{l}{LR} & \multicolumn{1}{l}{SVM} \\ \midrule
BC\_15               & 79.01                     & 87.57                   & 88.40                   & 75.97                      & \textbf{95.64}                   & \textbf{95.64}                    \\
Huong\_thom\_1         & 78.32                   & 93.50                   & 94.38                   & 81.53                    & 98.03                     & \textbf{98.18}                    \\
Nep\_87              & 86.91                   & 95.46                   & 95.46                   & 80.89                    & 97.25                   & \textbf{97.36}                    \\
Q\_5                 & 74.04                     & 85.71                   & 85.71                  & 77.67                    & \textbf{95.17}                   & 94.97                      \\
Thien\_uu\_8           & 90.21                   & 97.59                   & 97.44                   & 91.41                    & \textbf{98.80}                   & \textbf{98.80}                   \\
Xi\_23               & 77.50                   & 91.23                   & 92.55                   & 75.53                    & 95.33                   & \textbf{95.76}                   \\ \bottomrule
\end{tabular} 
\end{center}
\end{table}

\begin{table}[!th]

\caption{The {\red binary classification accuracy} (\%)   of proposed method based on features extracted from  ResNet-50  block 2 of Stage 5. The best results are in bold. \label{Resnet50_ML_block2}}
\begin{center}
\begin{tabular}{@{}lrrrrrr@{}}
\toprule
 Variety& \multicolumn{1}{l}{DT} & \multicolumn{1}{l}{ET} & \multicolumn{1}{l}{RF} & \multicolumn{1}{l}{KNN} & \multicolumn{1}{l}{LR} & \multicolumn{1}{l}{SVM} \\ \midrule
BC\_15               & 77.61                   & 84.77                   & 85.84                   & 75.55                    & 95.31                   & \textbf{95.72}                    \\
Huong\_thom\_1         & 78.10                   & 91.75                   & 92.26                   & 81.75                    & 97.37                   & \textbf{97.66}                    \\
Nep\_87              & 85.11                   & 94.51                   & 94.30                   & 81.84                    & \textbf{97.04}                     & 96.94                    \\
Q\_5                 & 72.84                   & 82.90                   & 82.80                  & 78.37                   & 93.36                  & \textbf{93.96}                   \\
Thien\_uu\_8           & 87.50                   & 96.99                  & 96.69                  & 93.07                   & 98.49                  & \textbf{98.80}                   \\
Xi\_23               & 71.37                  & 86.56                  & 87.07                  & 77.94                   & 95.03                  & \textbf{95.11}                   \\ \bottomrule
\end{tabular} 
\end{center}
\end{table}

\begin{table}[!th]
\caption{The {\red binary classification accuracy}  of proposed methods based on features extracted from ResNet-50 block 3 of Stage 5. The best results are in bold. \label{Resnet50_ML_block3}}
\begin{center}
    
\begin{tabular}{@{}lrrrrrr@{}}
\toprule
Variety & \multicolumn{1}{l}{DT} & \multicolumn{1}{l}{ET} & \multicolumn{1}{l}{RF} & \multicolumn{1}{l}{KNN} & \multicolumn{1}{l}{LR} & \multicolumn{1}{l}{SVM} \\ \midrule
BC\_15               & 74.32                  & 82.96                  & 82.88                  & 78.77                   & 94.32                  & \textbf{94.65}                  \\
Huong\_thom\_1         & 75.04                  & 90.73                  & 90.88                  & 80.88                   & 95.91                  & \textbf{96.06}                   \\
Nep\_87              & 83.74                  & 93.24                  & 93.24                  & 82.89                   & 96.83                  & \textbf{96.94}                   \\
Q\_5                 & 72.13                  & 81.39                  & 81.69                  & 76.96                   & 91.15                  & \textbf{91.25}                   \\
Thien\_uu\_8           & 87.05                  & 95.63                  & 95.93                  & 91.42                   & 97.89                  & \textbf{98.19}                   \\
Xi\_23               & 73.27                  & 84.15                  & 85.76                  & 78.6                    & \textbf{93.65}                  & 93.43                   \\ \bottomrule
\end{tabular} 
\end{center}
\end{table}

\section{Discussion}
 To improve the performance of the purity identification of rice seeds, we have suggested a hybrid machine learning-based model. Compared with existing methods solving the same task in the literature, the advantages of this study are as follows.
\begin{itemize}
\item   Firstly, the hybrid model combines two components, a CNN structure for extracting features from raw images and a machine learning algorithm for classifying the images. In the hybrid model, unlike most of the existing studies, we used the features extracted from not only the last block but also several previous blocks of the CNN structure. This allows to preservation of important features of the raw data that could be dismissed when compacting the features from block to block. To the best of our knowledge, this is the first time the technique has been considered to determine the purity of rice seeds. The obtained results show that our proposed strategy has led to a significant improvement in model performance.
  \item Secondly, in the experiment, we followed a practical setting for the task of identifying the rice seed purity. Theoretically, the multi-class classification setting in some previous studies may be more difficult and require models with higher complexity, however, it may not be suitable for the main goal of the task. Indeed, to determine if rice seed samples are pure, one must recognize whether it is mixed with other rice seed varieties, which means performing a binary classification problem. While the direct application of several CNN architectures did not result in a good performance, our proposed hybrid model can handle well the problem. 
  \item {\red Finally,  the proposed hybrid model outperforms the traditional ML method suggested in  \citet{hong2015comparative}. Based on the hand-crafted features, the ML models only achieved an accuracy of around 90\% in identifying rice seeds (in fact, the accuracy of most of the ML models was well below 90\%,  only a few ones approached this accuracy depending on the type of rice seeds).  Meanwhile, the hybrid model attained much higher performance. For example, by using the features extracted from the right blocks of VGG or ResNet-50, the LR  and SVM models led to an accuracy of above 95\% for all the rice seed types in the experiment. For some types of rice seeds like Huong\_thom\_1 and Thien\_uu\_8, the SVM model even reached an accuracy of up to about 99\% (see Table 3).  This shows the benefit of the proposed hybrid models in the study.   }
\end{itemize}

Regardless of the several advantages aforementioned, there are still certain limitations to the study. For example, due to the use of features extracted from the first blocks of the CNN architectures which are large, it costs a significant amount of time to train the machine learning models. Therefore, it would be remarkable progress if one could find other algorithms that reduce the training time of the model while still maintaining performance. Another limitation of the study is the use of only two CNN structures for the feature extraction. Although they provided an expected performance in the studies, it is not sure that they are the best. Moreover, the efficiency of a model depends also on the dataset, and there is no optimal model for all the datasets. It could be more practical if the performance of more deep learning structures has been investigated and a comparison between these structures corresponding to different datasets has been made to provide useful information for practitioners in appropriate choosing models.  Finally, deep learning models like VGG16 and Resnet50 are often considered black boxes as it is hard to interpret and understand the reasons for which the decisions are made. A deep insight into these deep learning structures is important to ensure that the designed models are understandable and reliable. One can consider these problems for further research. 



\section{Conclusion \label{sec: Conclude}}
We have proposed a novel method for identifying rice seed purity in this study. The main idea is to exploit the effective ability of deep learning models in extracting important features of rice seed images and the power of machine learning algorithms in classifying these features. In particular, features extracted from different blocks of two well-known deep  CNN architectures, namely VGG16 and ResNet-50, were fed to the machine learning algorithms. The experiments have been set up relying on consultation with experts in the field and practical implementation of the rice seed identification.  The obtained results showed that the proposed hybrid method performs better than the use of deep CNN models or machine learning models only, in which the combination of features extracted from block 3 of VGG16 and the SVM algorithm provides the best performance, achieving the highest accuracy of 99\%.  
The proposed method is then a base for designing automatic systems identifying rice seed purity, meeting the strict requirements in practice.

{\red\section*{Data availability statement}

The data used in the experiments of this study have not been submitted to publicly available repositories. Data will be made available from the corresponding author upon reasonable request. 

\section*{Ethics declarations}
Review and/or approval by an ethics committee was not needed for this study because [This work does not address the ethical considerations of animal, cell, and human experimentation.].}

\bibliographystyle{elsarticle-num}

\end{document}